\documentclass[conference]{IEEEtran}
\IEEEoverridecommandlockouts
\usepackage{graphicx}
\usepackage{amsmath,amssymb,amsfonts}
\usepackage{subcaption}
\usepackage{algorithmic}
\usepackage{cite}
\usepackage{threeparttable}  % 支持tablenotes
\usepackage{booktabs}       % 优化表格线（可选，更美观）
\usepackage{siunitx}        % 统一数值格式（可选）
\def\BibTeX{{\rm B\kern-.05em{\sc i\kern-.025em b}\kern-.08em
    T\kern-.1667em\lower.7ex\hbox{E}\kern-.125emX}}
\usepackage{float}
\IEEEoverridecommandlockouts                              % This command is only needed if 
\title{\LARGE \bf
Environment-Adaptive Solid-State LiDAR-Inertial Odometry
\thanks{This work was supported by the National Natural Science Foundation of China under Grants No. 62303083, U23A20318, 62221005, and 62441601, in part by the Science and Technology Research Program of Chongqing Municipal Education Commission (Grant No. KJQN202300626), and in part by the Science and Technology Innovation Key R\&D Program of Chongqing (Grant No. CSTB2023TIAD-STX0016).}
}

\begin{document}

\author{

\IEEEauthorblockN{1\textsuperscript{st} Zhi Zhang}
\IEEEauthorblockA{\textit{School of Automation} \\
\textit{Chongqing University of Posts and Telecommunications}\\
Chongqing, China \\
s230301076@stu.cqupt.edu.cn}
\and
\IEEEauthorblockN{2\textsuperscript{nd} Chalermchon Satirapod}
\IEEEauthorblockA{\textit{Department of Survey Engineering, Faculty of Engineering} \\
\textit{Chulalongkorn University}\\
Bangkok, Thailand \\
chalermchon.s@chula.ac.th}
\and
\IEEEauthorblockN{3\textsuperscript{rd} Bingtao Ma}
\IEEEauthorblockA{\textit{School of Cyberspace Security} \\
\textit{Hangzhou Dianzi University}\\
Hangzhou, China \\
mabingtao93@gmail.com}
\and
\IEEEauthorblockN{4\textsuperscript{th} Changjun Gu$^{*}$}
\IEEEauthorblockA{\textit{School of Automation} \\
\textit{Chongqing University of Posts and Telecommunications}\\
Chongqing, China \\
gucj@cqupt.edu.cn}
}
\def\IEEEtitletopspace{1pt}
\maketitle
\vspace{-1.0\baselineskip}
\thispagestyle{empty}
\pagestyle{empty}
%%%%%%%%%%%%%%%%%%%%%%%%%%%%%%%%%%%%%%%%%%%%%%%%%%%%%%%%%%%%%%%%%%%%%%%%%%%%%%%%
\begin{abstract}
Solid-state LiDAR-inertial SLAM has attracted significant attention due to its advantages in speed and robustness. However, achieving accurate mapping in extreme environments remains challenging due to severe geometric degeneracy and unreliable observations, which often lead to ill-conditioned optimization and map inconsistencies. To address these challenges, we propose an environment-adaptive solid-state LiDAR–inertial odometry that integrates local normal-vector constraints with degeneracy-aware map maintenance to enhance localization accuracy. Specifically, we introduce local normal-vector constraints to improve the stability of state estimation, effectively suppressing localization drift in degenerate scenarios. Furthermore, we design a degeneration-guided map update strategy to improve map precision. Benefiting from the refined map representation, localization accuracy is further enhanced in subsequent estimation. Experimental results demonstrate that the proposed method achieves superior mapping accuracy and robustness in extreme and perceptually degraded environments, with an average RMSE reduction of up to $12.8\%$ compared to the baseline method.

\end{abstract}
%%%%%%%%%%%%%%%%%%%%%%%%%%%%%%%%%%%%%%%%%%%%%%%%%%%%%%%%%%%%%%%%%%%%%%%%%%%%%%%%
\section{INTRODUCTION}

Simultaneous localization and mapping (SLAM) aims to simultaneously estimate sensor poses and incrementally construct a map, serving as a fundamental technology for robot navigation and autonomous driving \cite{jung2023helipr, gu2025information}. Existing methods can be broadly classified into visual SLAM (V-SLAM) and LiDAR-based SLAM (L-SLAM) \cite{wang2025sef}. V-SLAM achieves strong performance in structured environments; however, its robustness is sensitive to illumination variations \cite{chen2022overview, gu2025mis}. In contrast, L-SLAM provides accurate and illumination-invariant depth measurements, making it more reliable in large-scale and challenging environments \cite{zikos2011lslam, zhang2024inc}. In this paper, we focus on L-SLAM in degraded environments.

\begin{figure}[t]
    \centering
    \includegraphics[width=\linewidth]{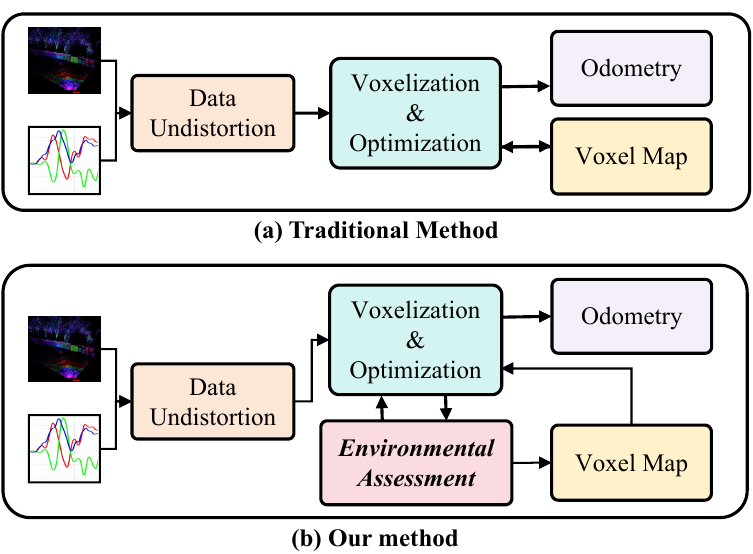}
    \caption{(a) Conventional method, where all measurements are directly integrated into the voxel map. (b) Our method incorporates environmental assessment into the optimization process by introducing environmental constraints and degeneracy evaluation to enhance localization accuracy.}
    \label{motivation}
    % \vspace{-15mm}
    \vspace{-0.2cm}
\end{figure}

As a major branch of L-SLAM, solid-state LiDAR–inertial SLAM integrates solid-state LiDAR and an IMU, which is widely used for pose estimation. For instance, LOAM-Livox utilizes the non-repetitive scanning of solid-state LiDARs to increase the number of feature points \cite{lin2020loamlivox}. Similarly, FAST-LIO significantly improves localization accuracy by tightly coupling LiDAR and IMU data \cite{xu2021fast, sun2021data}. Meanwhile, recent works have focused on system robustness in challenging environments. Notably, Zhang et al. enhanced the stability of state estimation by introducing online degeneracy detection\cite{zhang2016degeneracy}.

Although recent works have explored enhanced constraints and map representations, fundamental challenges remain in extreme scenarios, mainly due to the following factors:
% \begin{itemize}
% \item 
% \textbf{Localization drift in degenerate scenarios:} Relying on point-to-plane metrics that inadequately capture environment-specific characteristics in perceptually degraded environments, leading to reduced localization accuracy.
% \item 
% \textbf{Neglecting the impact of historical maps on localization:} Most existing approaches focus on front-end processing and state estimation, neglecting the impact of map quality on subsequent localization accuracy. As a result, localization performance can degrade in challenging environments.
% \end{itemize}

 1) \textbf{Localization drift in degenerate scenarios:} Relying on point-to-plane metrics that inadequately capture environment-specific characteristics in perceptually degraded environments, leading to reduced localization accuracy.
 
 2) \textbf{Neglecting the impact of historical maps on localization:} Most existing approaches focus on front-end processing and state estimation, neglecting the impact of map quality on subsequent localization accuracy. As a result, localization performance can degrade in challenging environments.

To address these challenges, we propose an environment-adaptive solid-state LiDAR–inertial odometry that aims to enhance localization accuracy. Specifically, we first estimate local surface normals within regions of varying radii centered at the same point to construct local normal-vector constraints, which allow the method to adapt to local environmental characteristics. Furthermore, we design a degeneration-guided map update strategy that regulates voxel updates based on confidence, preventing low-quality measurements from corrupting the global map. Extensive experiments on challenging real-world datasets demonstrate that the proposed method achieves superior localization accuracy, robustness, and mapping consistency compared with existing methods. The main contributions of this paper are summarized as follows:
\begin{itemize}
\item 
We propose a method that augments conventional point-to-plane features with planar angle constraints, explicitly leveraging environmental information to enhance localization and reconstruction accuracy.
\item 
We propose a degeneracy-aware adaptive voxel map update strategy that enables confidence-aware voxel updates and improves the consistency of map representation.
\item
Extensive experiments on the Botanic Garden datasets demonstrate the superiority of the proposed method.
\end{itemize}

\section{Related Work}
In this section, we briefly outline the related works focused on LiDAR-inertial SLAM Methods and degeneracy-aware SLAM Methods. 
\subsection{Lidar-Inertial SLAM Methods}
Autonomous robots are commonly equipped with LiDAR and IMU sensors~\cite{wisth2022vilens}. Existing LiDAR–inertial SLAM methods can be broadly classified into feature-based and feature-free approaches. Feature-based methods generally follow the LOAM framework~\cite{zhang2014loam}, which extracts edge and planar features based on local surface curvature. Building upon this paradigm, LeGO-LOAM~\cite{shan2018lego} improves pose estimation by leveraging ground segmentation and a two-stage optimization strategy. Benefiting from the lightweight and low-cost nature of solid-state LiDAR, FAST-LIO~\cite{xu2021fast} tightly couples LiDAR features with IMU measurements using an iterated extended Kalman filter, enabling efficient deployment on resource-constrained platforms. To further enhance robustness, IGE-LIO~\cite{chen2024ige} incorporates intensity edge features in addition to conventional geometric features.

In contrast, feature-free methods directly operate on raw point clouds without explicit feature extraction. For instance, Fast-LIO2~\cite{xu2022fast} removes the feature extraction module entirely and achieves superior real-time performance by directly leveraging raw point clouds, particularly in sparse environments encountered by solid-state LiDAR.

\subsection{Degeneracy-Aware SLAM Methods}
In degraded environments, the localization accuracy of SLAM systems often deteriorates due to insufficient geometric constraints~\cite{chen2024p2d}. Quantifying environmental degeneracy in LiDAR–inertial systems is therefore crucial for improving localization robustness~\cite{zhang2021lidar}. Existing approaches can be broadly categorized into eigenvalue-based and eigenvalue ratio-based methods. Eigenvalue-based methods evaluate degeneracy by analyzing the eigenvalues of the Hessian matrix, where small eigenvalues indicate weakly constrained directions~\cite{zhang2016degeneracy}. However, these methods are sensitive to measurement noise and point density. To alleviate this issue, eigenvalue ratio-based methods measure degeneracy using ratios between eigenvalues, which are more robust to environmental variations~\cite{ebadi2021dare}.

In contrast to these approaches, we explicitly model environmental degeneracy within a LiDAR–inertial odometry and mapping framework. By introducing angle constraints to enhance geometric consistency and a degeneracy-aware voxel map update strategy to regulate map maintenance based on measurement reliability, our method achieves more accurate and robust localization performance.

\section{METHOD}

We propose a solid-state LiDAR–inertial localization method to improve robot localization accuracy in complex environments. Section~\ref{sectionB} introduces a normal-vector angle constraint to complement conventional point-to-plane constraints. Furthermore, Section~\ref{sectionC} presents a degeneracy-aware map update strategy that improves the robustness and accuracy of localization and mapping in complex environments.

\begin{figure*}[t]
    \centering
    \includegraphics[width=\linewidth]{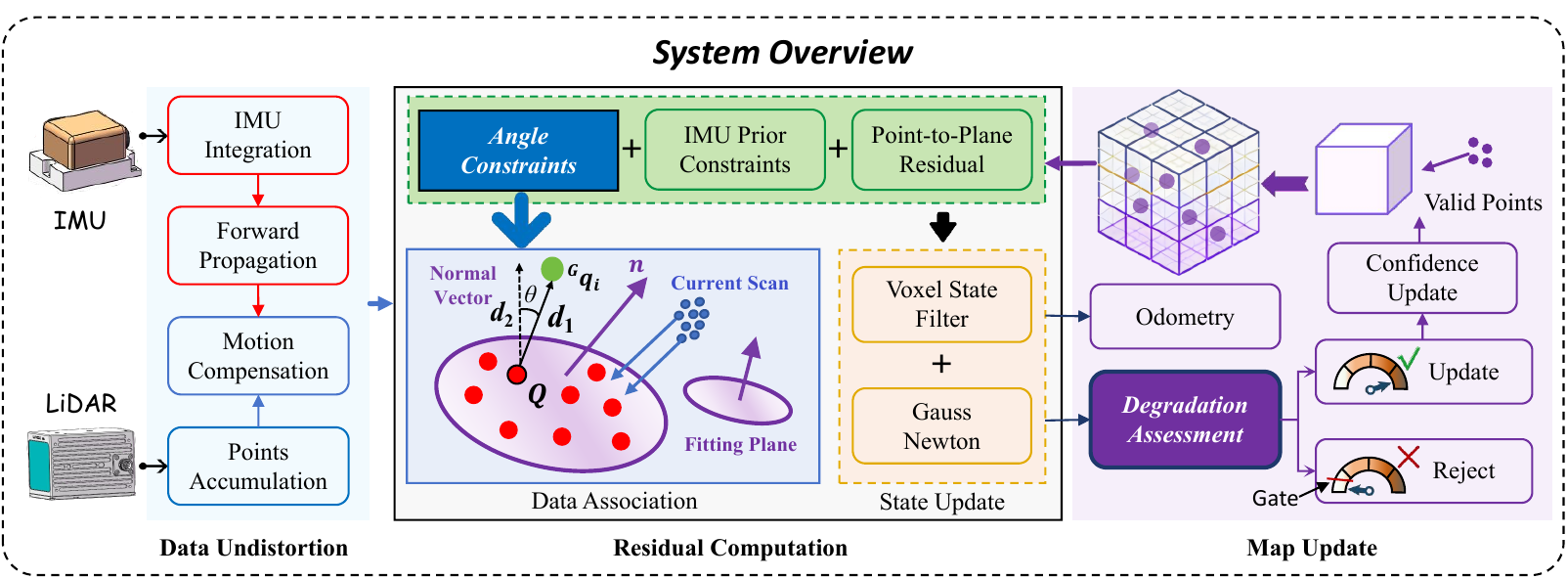}
    \caption{System overview of our proposed method. Firstly, motion distortion in the point cloud is removed by combining forward propagation of IMU measurements and backward propagation of LiDAR points. Then, three types of residuals are constructed to optimize pose estimation. Finally, environmental degeneracy is assessed, and the map is adaptively updated according to the confidence of the incoming point cloud measurements.}
    \label{fig:框架图}
    \vspace{-0.2cm}
\end{figure*}

\subsection{System Overview}\label{sectionA}
We define $I$ as the IMU frame and $L$ as the LiDAR frame. The system state can be defined as $X\doteq[^GR_I,^GP_I,^GV_I,b_\alpha,b_g]\in\mathcal{M}$, where $^GR_I$ and $^GP_I$ are the rotation and position of IMU in the global frame; $^{G}\nu_{I}$ is the linear velocity of IMU; $b_{\alpha}$ and $b_{g}$ describe three axis gyroscope bias and three axis accelerometer.

As shown in Fig. \ref{fig:框架图}, the proposed system framework focuses on residual computation from point cloud data and adaptive map updating. The pipeline begins with IMU pre-integration, which compensates for motion distortion in LiDAR scans while providing prior constraints propagated at the IMU frequency. The point cloud is then voxelized, and surface covariance is estimated within each voxel. In the optimization stage, GICP constraints~\cite{chen2024ig}, points matching residuals, and the proposed angle constraint are jointly employed to enhance geometric consistency. Meanwhile, the degree of environmental degeneracy is quantified to provide a reliability measure for subsequent map updating. Finally, the estimated pose is obtained via MAP optimization, and the voxel map is adaptively updated according to the degeneracy score to maintain map quality.

The MAP estimation for the $k$-th frame is formulated as
\begin{equation}
\min_{\boldsymbol{x}_{k}} \left\{ \sum_{i \in \mathcal{P}} \sum_{j \in \mathcal{V}} \left\| \boldsymbol{e}_{i,j} \right\|^2 + \left\| \boldsymbol{r}_{k}^{\mathrm{prior}} \right\|^2 \right\},
\label{MAP}
\end{equation}
where the total residual $\boldsymbol{e}_{i,j}$ consists of a point-to-plane residual $\boldsymbol{e}^{\mathrm{p2p}}$, a GICP residual $\boldsymbol{e}^{\mathrm{GICP}}$, and a normal-vector angle residual $\boldsymbol{e}^{\mathrm{ang}}$ (see Section~\ref{sectionB}). Here, $i$ denotes the $i$-th point in the point set $\mathcal{P}$, while $j$ indexes the voxel in the voxel set $\mathcal{V}$ to which the point belongs. The term $\boldsymbol{r}_{k}^{\mathrm{prior}}$ represents the prior constraint derived from IMU pre-integration \cite{chen2024ig}.

\subsection{Angle Constraints for Geometric Consistency}
\label{sectionB}

To enhance environmental adaptability, we introduce an angle constraint as a complement to conventional point-to-plane residuals. This design is motivated by the observation that surface normals on extended planar regions exhibit strong directional stability, making them robust geometric cues even under sensor noise and sparse measurements.

Consider a point $\boldsymbol{q}_i$ from the current LiDAR scan (in the local frame) and its corresponding point $\boldsymbol{Q}$ in the global map. Let $\boldsymbol{d}_1$ denote the vector connecting $\boldsymbol{Q}$ to the transformed LiDAR point in the global frame:
\begin{equation}
\boldsymbol{d}_1 = \boldsymbol{q}_i^G - \boldsymbol{Q}
= \boldsymbol{R} \left( \boldsymbol{q}_i^L - \boldsymbol{q}^L \right),
\label{d1}
\end{equation}
where $\boldsymbol{R}$ denotes the rotation from the LiDAR frame to the global frame, and $\boldsymbol{q}^L$ denotes the representation of $\boldsymbol{Q}$ in the LiDAR frame. 

To quantify angular consistency, we normalize $\boldsymbol{d}_1$ and compare it with the reference normal direction $\boldsymbol{d}_2$ of the fitted plane. The angle residual is defined as
\begin{equation}
e_\theta = 1 - \frac{\boldsymbol{d}_1}{\left\|\boldsymbol{d}_1\right\|} \cdot \boldsymbol{d}_2 .
\label{e_theta}
\end{equation}
For convenience, we simplify the notation of the normalized direction vector as
\begin{equation}
\boldsymbol{u} = \frac{\boldsymbol{d}_1}{\left\|\boldsymbol{d}_1\right\|}
= \frac{\boldsymbol{R}\,\Delta \boldsymbol{q}}{\left\|\boldsymbol{R}\,\Delta \boldsymbol{q}\right\|},
\label{u}
\end{equation}
where $\Delta \boldsymbol{q} = \boldsymbol{q}_i^L - \boldsymbol{q}^L$.

Then, we interpret the small angular differences between normal vectors as rotational perturbations. Owing to the limited angular variation on planar surfaces, the effect of such perturbations on the angle residual can be approximated by a first-order model, which allows the residual to be efficiently integrated into the optimization process. Under this assumption, the Jacobian of the normalized vector $\boldsymbol{u}$ with respect to $\boldsymbol{d}_1$ is derived as
\begin{equation}
\frac{\partial \boldsymbol{u}}{\partial \boldsymbol{d}_1}
=
\frac{1}{\left\|\boldsymbol{d}_1\right\|}
\left(
\boldsymbol{I} - \boldsymbol{u}\boldsymbol{u}^\top
\right).
\label{u/d}
\end{equation}

Meanwhile, the variation of the rotated vector with respect to the rotation perturbation $\delta \boldsymbol{\theta}$ can be expressed as
\begin{equation}
\frac{\partial \left( \boldsymbol{R}\,\Delta \boldsymbol{q} \right)}
{\partial \delta \boldsymbol{\theta}}
=
-\,\boldsymbol{R} \left[ \Delta \boldsymbol{q} \right]_\times,
\label{Rq/theta}
\end{equation}
where $[\cdot]_\times$ denotes the skew-symmetric matrix operator.

By applying the chain rule to combine the above derivatives, the Jacobian of the angle residual with respect to the rotational perturbation is finally obtained as
\begin{equation}
\frac{\partial e_\theta}{\partial \delta \boldsymbol{\theta}}
=
-\,\boldsymbol{d}_2^\top
\left(
\boldsymbol{I} - \boldsymbol{u}\boldsymbol{u}^\top
\right)
\frac{1}{\left\|\boldsymbol{d}_1\right\|}
\left(
-\,\boldsymbol{R} [\Delta \boldsymbol{q}]_\times
\right).
\label{e_ang}
\end{equation}

\subsection{Degeneracy-Aware Voxel Map Update Strategy}\label{sectionC}
Although the proposed angle constraints improve geometric consistency and pose observability on planar structures, their effectiveness degrades in highly degenerate scenes with weak constraints. In such cases, indiscriminate map updates may propagate uncertainty and degrade map quality. To mitigate this issue, we introduce a degeneracy-aware adaptive voxel map update strategy to ensure stable and reliable map construction.

\subsubsection{Degeneracy Assessment from Short-Term Data}
In MAP-based state estimation, environmental degeneracy manifests as poor conditioning of the Hessian matrix. When geometric constraints are insufficient, the Hessian becomes ill-conditioned or rank-deficient, leading to unstable pose updates. Based on this observation, we evaluate degeneracy by analyzing the accumulated Hessian matrix obtained during the optimization process, without introducing additional heuristic indicators.

Specifically, after linearizing all residuals in the MAP formulation, including both point-to-plane distance residuals and normal-vector angle residuals, the approximate Hessian matrix is accumulated as
\begin{equation}
\mathbf{H} = \sum \mathbf{J}^\top \mathbf{J},
\label{H}
\end{equation}
where $\mathbf{J}$ denotes the Jacobian of each residual with respect to the state variables. To explicitly characterize the observability of rotational and translational components, we partition the Hessian matrix into rotational and translational sub-blocks:
\begin{equation}
\mathbf{H}_d = \begin{bmatrix}
\mathbf{H}_{rr} & \mathbf{H}_{rt} \\
\mathbf{H}_{tr} & \mathbf{H}_{tt}
\end{bmatrix},
\label{Hd分块矩阵}
\end{equation}
where $\mathbf{H}_{rr} \in \mathbb{R}^{3 \times 3}$ and $\mathbf{H}_{tt} \in \mathbb{R}^{3 \times 3}$ correspond to rotation and translation, respectively, and can be used to analyze the degeneracy of the environment during sensor motion.

Then we compute the condition numbers of the sub-matrices $\mathbf{H}_{rr}$ and $\mathbf{H}_{tt}$ through Singular Value Decomposition (SVD). For each sub-matrix, the condition number $\kappa$ is calculated as the ratio of the largest to smallest singular values ($\lambda_{max}$/$\lambda_{min}$). Following the analysis in \cite{belsley2005regression}, higher $\kappa$ values indicate greater system instability, weak constraints, and potential degeneracy along the corresponding degrees of freedom. Then we define two degeneracy factors:
\begin{equation}
\kappa_r = \frac{\lambda_{r,\text{max}}}{\max(\lambda_{r,\text{min}}, \epsilon)}, \quad \kappa_t = \frac{\lambda_{t,\text{max}}}{\max(\lambda_{t,\text{min}}, \epsilon)},
\label{k_r,k_t}
\end{equation}
where $\epsilon$ is a small constant preventing division by zero.

To obtain a normalized and numerically stable degeneracy indicator, the condition numbers are further mapped into bounded observability scores:
\begin{equation}
\begin{aligned}
s_r &= 1 - \exp\left(-\left(1 - \frac{1}{\max(\kappa_r, 1)}\right)\right), \\
s_t &= 1 - \exp\left(-\left(1 - \frac{1}{\max(\kappa_t, 1)}\right)\right).
\end{aligned}
\label{s退化分数}
\end{equation}

These scores decrease monotonically as the system becomes more ill-conditioned, providing a smooth measure of structural degeneracy.
Finally, the rotational and translational degeneracy scores are fused to form an overall structural stability score:
\begin{equation}
s_{\text{struct}} = \alpha s_r + (1 - \alpha) s_t,
\label{s_struct}
\end{equation}
where $\alpha$ controls the relative importance between rotational and translational constraints.

Meanwhile, we introduce an angle-based stability score derived from the normal-angle residuals. First, we compute the mean angular residual over all valid correspondences:
\begin{equation}
\bar{e}_\theta = \frac{1}{N} \sum_{i=1}^{N} e_{\theta,i}.
\label{bar_theta}
\end{equation}

Then, a bounded angle-based stability score is computed as
\begin{equation}
s_{\text{angle}} = \exp\left(-\bar{e}_\theta\right),
\label{s_angle}
\end{equation}
where $s_{\text{angle}} \in (0, 1]$. Then, the structural stability score obtained from Hessian analysis and the angle-based stability score are jointly fused to form a unified degeneration score:
\begin{equation}
s_{\text{deg}} = \gamma s_{\text{struct}} + (1 - \gamma) s_{\text{angle}}.
\label{s_deg}
\end{equation}

This combined score captures both global observability and local geometric consistency. As the Hessian matrix is already accumulated during MAP optimization, the proposed assessment incurs negligible computational overhead and is seamlessly integrated into the estimation process. The resulting degeneration score is then used to guide the degeneracy-aware voxel map update strategy (see Section~\ref{map_update}).

\subsubsection{Degeneracy-Guided Voxel Update Strategy}
\label{map_update}
To mitigate the adverse effects of degenerate measurements on long-term map consistency, we propose a degeneracy-guided voxel update strategy that adaptively regulates map maintenance according to the estimated score $s_{\text{deg}}$. Instead of blindly accumulating incoming points, the voxel update strategy is dynamically adjusted according to both the current point quality and the historical confidence of each voxel.

At the frame level, map update is completely disabled when severe degradation is detected:
\begin{equation}
s_{\text{deg}} < \tau_{\text{global}},
\end{equation}
where $\tau_{\text{global}}$ is the global gating. If the score falls below the predefined threshold, the current scan is excluded from map integration, thus preventing large pose uncertainty from propagating into the voxel representation.

For valid frames, voxel creation and replacement are regulated by the degeneracy score, allowing updates only when incoming measurements are sufficiently reliable. 
Specifically, the confidence of a newly created voxel is initialized by the degeneracy score of its first integrated point, i.e., $q_{v}=s_{d}$. The voxel confidence is then updated according to the following rules:
% \begin{itemize}
% \item 
% In the event of a hash collision, the existing voxel is removed and replaced by a newly created voxel.
% \item 
% When additional point cloud observations are incorporated, the voxel confidence is updated using the following formulation:
% \begin{equation}
% q_v^{'} =\beta q_v + (1 - \beta) s_{\text{deg}}
% \label{qv}
% \end{equation}
% \end{itemize}

1) In the event of a hash collision, the existing voxel is removed and replaced by a newly created voxel.

2) When additional point cloud observations are incorporated, the voxel confidence is updated using the following formulation:
\begin{equation}
q_v^{'} =\beta q_v + (1 - \beta) s_{\text{deg}},
\label{qv}
\end{equation}
where $\beta$ is an empirically determined weighting coefficient that balances the influence of newly added point clouds on the voxel confidence.

Overall, by incorporating angular constraints, performing an environmental degeneracy assessment, and designing a degeneracy-guided map update strategy, the proposed method achieves high-accuracy localization in complex environments and significantly improves environmental adaptability.

\section{EXPERIMENTAL RESULTS}
In this section, we first introduce the experimental setup in Section \ref{实验设置}. Then, we validate our proposed method on the Botanic Garden dataset\cite{liu2024botanicgarden}. The experimental results show the effectiveness and advancement of our proposed method.

\subsection{Experimental Setup}
\label{实验设置}
\textbf{Evaluation Metrics:}
We utilize Absolute Pose Error (APE) with three metrics: Maximum Error, Mean Error, and Root Mean Squared Error (RMSE). All experimental results are obtained by running each method five consecutive times and reporting the average performance. And all experiments are conducted on a laptop, with hyperparameters kept constant to ensure reproducibility.

\textbf{Datasets:}
The Botanic Garden dataset is a high-quality multi-sensor robotic navigation dataset collected in a large-scale botanical garden using an all-terrain wheeled robot equipped with a time-synchronized solid-state LiDAR, IMU, and camera. The dataset covers diverse and challenging outdoor scenarios, including dense woods, riversides, narrow trails, bridges, and open meadows, providing rich geometric variations for evaluating localization and mapping robustness.

\begin{figure}[t]
    \centering
    \includegraphics[width=\linewidth]{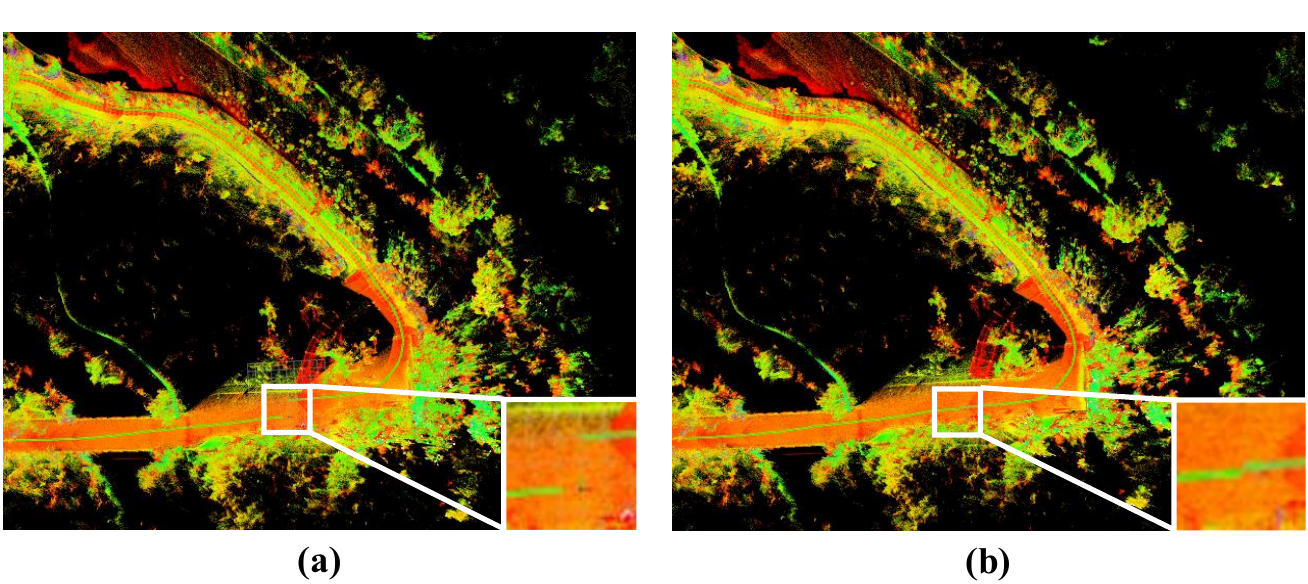}
    \caption{Qualitative comparison with iG-LIO on the Botanic Garden dataset: (a) shows the result obtained using the iG-LIO method;(b) shows the result obtained using the proposed full method.}
    \label{fig:对比图}
    \vspace{-0.2cm}
\end{figure}

\begin{figure*}[t]
    \centering
    \includegraphics[width=\linewidth]{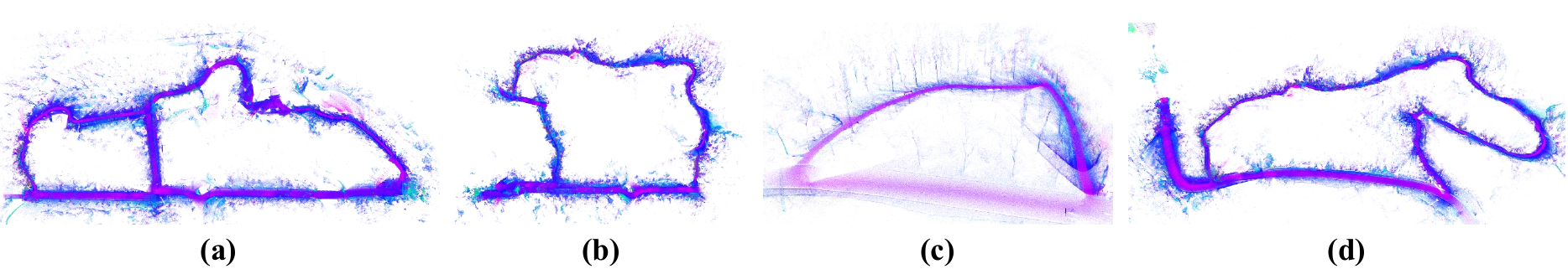}
    \caption{Reconstruction results constructed by the proposed method on the Botanic Garden dataset.}
    \label{fig:建图结果}
    \vspace{-0.2cm}
\end{figure*}

\subsection{Comparison Experiments}\label{对比实验}
Tables~\ref{tab:rmse}, \ref{tab:max_error}, and \ref{tab:mean_error} report the quantitative localization errors of different methods on the Botanic Garden dataset, including FAST-LIO, iG-LIO, and the proposed method.

From Tables~\ref{tab:rmse}–\ref{tab:mean_error}, FAST-LIO shows relatively large errors across all sequences, indicating limited robustness in geometrically degraded environments. iG-LIO improves localization accuracy through voxel-based GICP constraints, but its performance still degrades in challenging sequences. In contrast, the proposed method achieves the best or comparable results across all metrics. As shown in Table~\ref{tab:rmse}, the full system attains the lowest RMSE in most sequences, while the maximum and mean error statistics (Tables~\ref{tab:max_error} and~\ref{tab:mean_error}) further demonstrate its ability to suppress large deviations and produce more stable trajectories. These results confirm that the integration of angle constraints and confidence-aware map maintenance significantly enhances robustness and localization accuracy in perceptually degraded environments.

\begin{table}[!t]
\centering
\caption{RMSE of Different Methods on Botanic Garden Dataset (Unit: m)}
\label{tab:rmse}
\begin{threeparttable}
\begin{tabular}{l S[table-format=1.3] S[table-format=1.3] S[table-format=1.3] S[table-format=1.3]}
\toprule
Sequence & {FAST-LIO} & {iG-LIO} & {Ours (w/o Deg)} & {Ours (Full)} \\
\midrule
1018-00 & 1.512 & 0.072 & 0.072 & 0.072 \\
1018-01 & 1.325 & 0.641 & 0.532 & \textbf{0.504} \\
1018-02 & 1.468 & 0.196 & 0.192 & \textbf{0.177} \\
1018-03 & 0.875 & 0.166 & 0.170 & \textbf{0.166} \\
1018-04 & 1.392 & 1.218 & 0.840 & \textbf{0.806} \\
1018-05 & 0.422 & 0.324 & 0.286 & \textbf{0.272} \\
1018-06 & 1.092 & 0.765 & 0.684 & \textbf{0.678} \\
\bottomrule
\end{tabular}
\begin{tablenotes}[flushleft]
\footnotesize
\item \textit{Note}: w/o Deg denotes the variant without the proposed degeneracy-aware voxel map update strategy.
\end{tablenotes}
\end{threeparttable}
\end{table}

\begin{table}[!t]
\centering
\caption{Maximum Error of Different Methods on Botanic Garden Dataset (Unit: m)}
\label{tab:max_error}
\begin{threeparttable}
\begin{tabular}{l S[table-format=1.3] S[table-format=1.3] S[table-format=1.3] S[table-format=1.3]}
\toprule
Sequence & {FAST-LIO} & {iG-LIO} & {Ours (w/o Deg)} & {Ours (Full)} \\
\midrule
1018-00 & 2.491 & 0.174 & 0.158 & \textbf{0.152} \\
1018-01 & 2.485 & 1.528 & 1.236 & \textbf{1.078} \\
1018-02 & 3.064 & 0.434 & 0.454 & \textbf{0.420} \\
1018-03 & 1.751 & 0.332 & 0.336 & \textbf{0.320} \\
1018-04 & 3.382 & 2.878 & 2.501 & \textbf{1.942} \\
1018-05 & 0.980 & 0.630 & 0.534 & \textbf{0.514} \\
1018-06 & 2.046 & 1.558 & 1.424 & \textbf{1.386} \\
\bottomrule
\end{tabular}
\begin{tablenotes}[flushleft]
\footnotesize
\item \textit{Note}: w/o Deg denotes the variant without the proposed degeneracy-aware voxel map update strategy.
\end{tablenotes}
\end{threeparttable}
\end{table}

\begin{table}[!t]
\centering
\caption{Mean Error of Different Methods on Botanic Garden Dataset (Unit: m)}
\label{tab:mean_error}
\begin{threeparttable}
\begin{tabular}{l S[table-format=1.3] S[table-format=1.3] S[table-format=1.3] S[table-format=1.3]}
\toprule
Sequence & {FAST-LIO} & {iG-LIO} & {Ours (w/o Deg)} & {Ours (Full)} \\
\midrule
1018-00 & 1.364 & 0.070 & 0.068 & \textbf{0.066} \\
1018-01 & 1.165 & 0.556 & 0.466 & \textbf{0.438} \\
1018-02 & 1.270 & 0.288 & 0.168 & \textbf{0.161} \\
1018-03 & 0.771 & \textbf{0.154} & 0.162 & 0.156 \\
1018-04 & 1.158 & 1.011 & 0.726 & \textbf{0.652} \\
1018-05 & 0.340 & 0.270 & 0.246 & \textbf{0.234} \\
1018-06 & 0.912 & 0.654 & 0.592 & \textbf{0.590} \\
\bottomrule
\end{tabular}
\begin{tablenotes}[flushleft]
\footnotesize
\item \textit{Note}: w/o Deg denotes the variant without the proposed degeneracy-aware voxel map update strategy.
\end{tablenotes}
\end{threeparttable}
\vspace{-6pt}
\end{table}

\subsection{Ablation Study}\label{消融实验}
To evaluate the contribution of each component, we conduct an ablation study on the Botanic Garden dataset by progressively augmenting the baseline iG-LIO. Three configurations are considered: iG-LIO, Ours (w/o Deg), which incorporates the proposed normal-vector angle constraints without degeneracy-aware map updating, and Ours (Full), which further enables degeneracy-aware adaptive voxel updates.

Compared with iG-LIO, Ours (w/o Deg) achieves consistent improvements in RMSE, maximum error, and mean error in most sequences (Tables~\ref{tab:rmse}–\ref{tab:mean_error}), demonstrating that the proposed angle residual effectively enhances geometric consistency and pose observability, especially in planar-dominant scenes. With the degeneracy-aware map update strategy enabled, Ours (Full) further improves performance in challenging sequences such as 1018-01, 1018-04, and 1018-06. In particular, on sequence 1018-04, the maximum error is reduced to $1.942\mathrm{m}$, representing reductions of $32.5\%$ and $22.4\%$ compared to iG-LIO and Ours (w/o Deg), respectively.  Fig.~\ref{fig:建图结果} shows the resulting map, illustrating the ability of the proposed method to construct accurate and consistent maps in complex environments.
% These results confirm that the degeneracy-aware map update can effectively suppress low-quality measurements, thereby improving mapping robustness and map consistency.
% \begin{table}[htbp]

Overall, the proposed method consistently outperforms existing LIO baseline methods in terms of accuracy and robustness. The integration of angle constraints and degeneracy-aware voxel updates effectively improves pose observability and suppresses error accumulation in challenging environments, enabling stable localization and high-quality mapping.

\section{CONCLUSION}
In this paper, we propose an environment-adaptive solid-state LiDAR–inertial odometry and mapping framework to improve robustness in geometrically degenerate environments. By introducing normal-vector angle constraints, the proposed method enhances local geometric consistency and improves pose estimation accuracy. In addition, a degeneracy-aware voxel map update strategy is developed to adaptively regulate map updates based on measurement reliability, effectively mitigating the impact of low-quality observations. Experimental results on the Botanic Garden dataset demonstrate that the proposed approach consistently outperforms existing methods in terms of accuracy and robustness, particularly in challenging scenarios with limited or repetitive structures. Future work will investigate extending the proposed strategy to more complex and dynamic environments.

\bibliographystyle{IEEEtran} 
\bibliography{IEEEabrv,Ref}

\end{document}